\documentclass[journal]{IEEEtran}
\usepackage{graphicx}
\usepackage{dcolumn}
\usepackage{bm}
\usepackage{algorithm}
\usepackage{algpseudocode}
\usepackage{algorithmicx}
\usepackage{booktabs}
\usepackage{rotating}
\usepackage[table,xcdraw]{xcolor}
\usepackage{cite}
\begin{document}

\title{Transfer learning to model inertial confinement fusion experiments}

\author{
    \IEEEauthorblockN{K. D. Humbird\IEEEauthorrefmark{1}\IEEEauthorrefmark{2}, J. L. Peterson\IEEEauthorrefmark{1}, R. G. McClarren\IEEEauthorrefmark{3}} \\
    \IEEEauthorblockA{\IEEEauthorrefmark{1}Lawrence Livermore National Laboratory, 7000 East Ave, Livermore, CA 94550} \\
    \IEEEauthorblockA{\IEEEauthorrefmark{2}Department of Nuclear Engineering, Texas A \& M University, 3133 TAMU, College Station, TX 77843}
\IEEEauthorblockA{\IEEEauthorrefmark{3}Department of Aerospace and Mechanical Engineering, University of Notre Dame, 365 Fitzpatrick Hall, Notre Dame, IN 46556}
}

%

\maketitle
\date{\today}
\begin{abstract}
Inertial confinement fusion (ICF) experiments are designed using computer simulations that are approximations of reality, and therefore must be calibrated to accurately predict experimental observations. In this work, we propose a novel nonlinear technique for calibrating from simulations to experiments, or from low fidelity simulations to high fidelity simulations, via ``transfer learning''. Transfer learning is a commonly used technique in the machine learning community, in which models trained on one task are partially retrained to solve a separate, but related task, for which there is a limited quantity of data. We introduce the idea of hierarchical transfer learning, in which neural networks trained on low fidelity models are calibrated to high fidelity models, then to experimental data. This technique essentially bootstraps the calibration process, enabling the creation of models which predict high fidelity simulations or experiments with minimal computational cost. We apply this technique to a database of ICF simulations and experiments carried out at the Omega laser facility. Transfer learning with deep neural networks enables the creation of models that are more predictive of Omega experiments than simulations alone. The calibrated models accurately predict future Omega experiments, and are used to search for new, optimal implosion designs. 
\end{abstract}


\section{Introduction}
Many physical systems and experiments are designed using models -- analytical theories or computer simulations that attempt to take into account the various components of the system to determine the efficiency, performance, and reliability of a design. In applications where the physics of the system is well known, the models can be accurate depictions of reality, however when dealing with systems at extreme conditions (such as extremely high temperatures, pressures, and densities), the physics is not as well understood and the models are not always validated with experimental data. This is often the case in inertial confinement fusion (ICF), in which lasers are used to compress a small fuel capsules filled with deuterium and tritium to high density, temperature, and pressure in order to create conditions that are favorable for nuclear fusion reactions~\cite{Atzeni,Lindl}. The computer simulations that model ICF experiments are complex and involve a wide variety of physics models: radiation hydrodynamics, atomic physics, nuclear burn physics, laser and plasma physics, magnetic field effects, and more~\cite{hydra}. These codes are validated in certain regimes, but acquiring data at the extreme conditions reached in ICF experiments is challenging and expensive, thus the accuracy of the models away from the validation data is not well known. Furthermore, fully-integrated simulations which model everything from the laser beam propagation to the particle transport within the fuel capsule are extremely expensive to run in 3D with high resolution. Researchers often need to make many approximations, such as running the simulation in 2D with axi-symmetric constraints, in order to efficiently search the parameter space for promising experimental designs. 

When the simulations used to design new ICF experiments contain simplifying assumptions, it is expected that there will be discrepancies between the simulator prediction and what is observed in the experiment. A common statistical approach to correcting an inaccurate simulator is model calibration -- using experimental data to ``calibrate'' an inaccurate model to produce one which is more consistent with reality. Model calibration is a broadly-researched topic~\cite{bayescalib,bayescalib2, bayescalib1,mcclarren}, with one of the most popular techniques developed by Kennedy and O'Hagan~\cite{kennedyohagan}. In this approach, the true model is assumed to be an additive combination of a simulator, Gaussian distributed error due to measurement uncertainty, and an unknown discrepancy term which is learned using experimental data. The form of this discrepancy term is often specified by the user (for example, the user might chose a Gaussian process discrepancy with a particular kernel function) and the complexity is limited by the amount of experimental data that is available. Researchers have explored Bayesian calibration with discrepancy terms for ICF data~\cite{JimBayesian}; in this work, we propose an alternative calibration technique borrowed from the machine learning community, referred to as ``transfer learning'', for calibrating ICF models.

Traditional machine learning models gain knowledge by observing large quantities of labeled data. For example, if the task is to classify photos of animals, a model will need to be exposed to millions of labeled images of all the animals it is expected to classify, in a variety of different scenarios, colors, perspectives, etc, in order to classify the animals correctly. Supervised learning tasks are straightforward to solve when large quantities of data are available, however many of these techniques break down when limited to small sets of labeled data. 

Transfer learning is an alternative learning technique that can help overcome the challenge of training on small datasets. Transfer learning is a method for using knowledge gained while solving one task, and applying it to a different, but related task. This approach is most commonly used for image classification~\cite{TLimages, TLimages1, TLimages2}, for which there are many large databases of labeled images~\cite{imagenet,pascal,caltech} and several pre-trained neural network models that are available for download, such as AlexNet~\cite{alexnet} and Inception~\cite{inception}. These neural networks have been studied extensively, and they appear to learn how to recognize images in a logical series of steps as you traverse hidden layers in the models. First, the networks often search for edges in the images, then for simple geometric patterns, and eventually begin to recognize specific characteristics, such as eyes, arms, ears, etc~\cite{inception}. In general, the deeper in the network you go, the finer the details the network appears to focus on. It therefore seems logical to expect that for a neural network trained on any image dataset, the first several hidden layers are essentially the same -- they learn about features common to all images. One can thus take a pre-trained neural network, freeze the first several layers of weights, and focus on re-training only the last few layers on a new image dataset to learn how to appropriately classify this new set of images. The old frozen layers are where the network learns to ``see''; the last few are where the network learns to ``recognize'' specific types of images. Transfer learning is the process of taking a network trained on a large dataset, freezing several layers of the network, then retraining the unfrozen layers on a different, often smaller dataset. As an example relevant to ICF, researchers at the National Ignition Facility (NIF)~\cite{NIFMiller} have used transfer learning to classify images of different types of damage that occur on the optics at NIF. There are not enough labeled optics images to train a network from scratch, but transfer learning with a network pre-trained on ImageNet~\cite{imagenet} produces models which classify optics damage with over 98\% accuracy. This methodology has enabled the group to automate their damage inspection by letting the network process the images of optical components, rather than having an optics expert inspect each image manually~\cite{NIFoptics}. 

\begin{figure}
\begin{center}
		\includegraphics[width=0.45\textwidth]{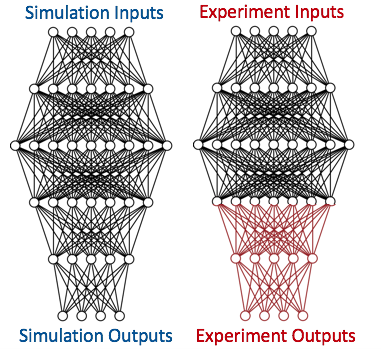}
		\caption{To transfer learn from simulations to experiments, the first three layers of the simulation-based network are frozen, and the remaining two layers are available for retraining with the experimental data.}
		\label{fig:tl2}
\end{center}
\end{figure}

In ICF it is often the case that we do not have enough experimental measurements to train a machine learning model on the experimental data alone. We are not interested in image classification for our implosions, so traditional transfer learning using a pre-trained open-source model is not appropriate. However, we do have the ability to create massive databases of ICF simulations~\cite{langer,Peterson2017}, which we suspect are a good reflection of reality, but need to be tuned to be more consistent with experiments. We therefore propose the use of transfer learning as a non-parametric approach for calibrating a simulation-based neural network to experiments. The general concept is illustrated in Figure \ref{fig:tl2}: train a feed-forward neural network on simulation data to relate simulation inputs to observable outputs. Freeze many of the layers of the neural network, but leave some open for re-training. Retrain these available weights using the sparse set of experimental data for which the inputs and output observables are known. 

We will test the feasibility of using transfer learning to produce neural networks that accurately predict the outcomes of ICF experiments carried out at the Omega laser facility~\cite{omega} at the Laboratory for Laser Energetics (LLE). In section~\ref{sec:tl1} we will introduce a proposed hierarchical approach to transfer learning for numerical simulation and experimental data. In section~\ref{sec:tl2}, we will compare standard and hierarchical transfer learning for the Omega dataset, and use the transfer learned models to study the discrepancies between the Omega simulations and experiments in section~\ref{sec:tl3}.

\section{Hierarchical transfer learning}\label{sec:tl1}
Computer simulations of complex physical systems are often modeled at varying levels of fidelity. Fast, low fidelity models are used to explore vast design spaces for optimal settings, and expensive high fidelity models might be used in interesting regions of design space to compute predictions of planned experiments. The high fidelity simulations are often more accurate and reliable than the fast, approximate models, but the expense of running the simulation often prevents their use in large parameter scans. It might be possible to create models that emulate high fidelity simulations with reduced computational cost by using transfer learning. For example, a model trained on a dense set of 1D simulations could be calibrated to a sparse set of 2D simulations that fill the same design space. Rather than running a dense set of 2D simulations to train a 2D surrogate model, an equally accurate surrogate might be obtained by transfer learning from 1D to 2D with a relatively small number of 2D simulations, saving substantial computational resources. Furthermore, this 2D-calibrated model can be subsequently calibrated to experimental data. If the 2D model is a better reflection of reality than the 1D model, the transfer learning step between 2D and the experiment should be easier than jumping from 1D directly to the experimental data. We refer to this technique of transfer learning from low to subsequently higher fidelity simulations to the experimental data as ``hierarchical transfer learning''. 

To demonstrate the utility of hierarchical transfer learning, consider the simple function: 

\begin{equation}\label{eq:1}
f(a,x)= xe^{ax},
\end{equation} 

where $x$ and $a$ are random variables; $x$ between [-1,1] and $a$ between [0,1]. This expression will be the ``experiment'' or true function. We also have a low and high fidelity approximation of the experiment:

\begin{equation}\label{eq:2}
f_{low}(a,x) = x,
\end{equation}

\begin{equation}\label{eq:3}
f_{high}(a,x) = x+ax^2,
\end{equation}

where these are the first (low fidelity) and second (high fidelity) order Taylor expansions for the true function. This problem is used to study the benefits of hierarchical transfer learning; specifically to see if stepping through the hierarchy results in better models than calibrating directly from low fidelity to experimental data.

The feed-forward neural networks used in this study are designed with the algorithm ``deep jointly-informed neural networks'' (DJINN)~\cite{djinn}. For this comparison, DJINN models with 3 hidden layers are trained to map from $(x,a)$ to $f(x,a)$ or one of the approximations in Equations~\ref{eq:2}-\ref{eq:3}. First we compute the average explained variance score (averaged over 5 random training/testing data splits of 80/20\%) for DJINN models trained on experiments alone; this is the baseline to which we will compare various transfer learning techniques, as we do not expect them to exceed the performance of a DJINN model trained purely on experimental data. Next, we transfer learn from high fidelity simulations to experiments, then from low fidelity simulations to experiments. Finally, we transfer learn from low to high fidelity, and then to experiments to test the hierarchical approach. The results are summarized in Table \ref{table:ex}; the neural network hyper-parameters are noted in Table \ref{table:exhp} and are kept the same for all of the models.

\begin{table}[]
\caption{Hyper-parameters for original model and transfer learning for the Taylor expansion example.}
\label{table:exhp}
\begin{center}\begin{tabular}{ll}
\hline
\multicolumn{2}{c}{Original Model Parameters} \\ \hline
\multicolumn{1}{l|}{Number of models} & 5 \\
\multicolumn{1}{l|}{Hidden layer widths} & \begin{tabular}[c]{@{}l@{}}4-8-14; 4-7-15, 4-8-14; \\ 4-7-15; 4-7-14\end{tabular} \\
\multicolumn{1}{l|}{Learning rate} & 0.004 \\
\multicolumn{1}{l|}{Batch size} & 50 \\
\multicolumn{1}{l|}{Epochs} & 300 \\ \hline
\multicolumn{2}{c}{Transfer Learning Parameters} \\ \hline
\multicolumn{1}{l|}{Retrained layers} & Final 2 \\
\multicolumn{1}{l|}{Learning rate} & 0.0001 \\
\multicolumn{1}{l|}{Batch size} & 1 \\
\multicolumn{1}{l|}{Epochs} & 300
\end{tabular}\end{center}
\end{table}

\begin{table*}[]
\caption{Comparison of hierarchical and one-step transfer learning (TL) to direct modeling of experiments. Low/high fi. indicates data produced with the low/high fidelity simulations, respectively. Exp. indicates ``experimental'' data produced with the analytic expression in Eq.~\ref{eq:1}.}
\label{table:ex}\begin{center}
\begin{tabular}{c|c|c}
\hline
Model                                                                                                         & \begin{tabular}[c]{@{}c@{}}Mean$\pm$SD \\ Explained Variance\end{tabular} & p-value \\ \hline
Train with 100 exp.                                                                                           & 0.994 $\pm$ 0.004                                                         & -       \\ \hline
\begin{tabular}[c]{@{}c@{}}Train with 100 high fi.; \\ TL with 25 exp.\end{tabular}                          & 0.994 $\pm$ 0.006                                                         & 0.957   \\ \hline
\begin{tabular}[c]{@{}c@{}}Train with 100 low fi.; \\ TL with 25 exp.\end{tabular}                           & 0.954 $\pm$ 0.041                                                         & 0.016   \\ \hline
\begin{tabular}[c]{@{}c@{}}Train with 100 low fi.; \\ TL with 50 high fi. \\ + TL with 25 exp.\end{tabular} & 0.981 $\pm$ 0.025                                                         & 0.279   \\ \hline
\end{tabular}\end{center}
\end{table*}

For this simple example, there is not a statistically significant difference between a model trained exclusively on a large dataset of experiments, models that are trained on high fidelity simulations and calibrated to experiments, and models that are hierarchically calibrated. However, these three models are statistically significantly better than the model which is calibrated directly from low fidelity to the experiments, illustrating that there is an advantage of informing the model of high fidelity data prior to experimental calibration. In order to make an accurate emulator of the experiments, the cost of each of these routes should be the determining factor in which method to choose; however it is expected in most cases that the hierarchical method, which requires the least number of experiments and/or high fidelity simulations, is the least expensive approach. For this simple example the computational cost is negligible, but for applications in which complex multi-physics systems are being modeled, the computational cost difference between low and high fidelity simulations can be substantial; for such systems experiments are also often costly and limited in number. 

The choice of performing the hierarchical transfer learning with 50 high fidelity simulations and 25 experiments in Table \ref{table:ex} is arbitrary. To determine the minimum number of high fidelity simulations and experiments that are necessary to produce a model that is not significantly different than the baseline (experiment only) surrogate, we can compute the mean explained variance score as the dataset sizes are varied. 

First, we determine the minimum number of high-fidelity points that are required to produce a transfer-learned model that is of similar performance to a model trained exclusively on 100 high fidelity simulations. Then, we determine how many experiments are needed to recalibrate this model to be on par with the experiment-only baseline model. As in the models from Table \ref{table:ex}, the number of points in the dataset includes training and testing data, which are split into 80/20\% sets. The transfer learning parameters are held constant as recorded in Table \ref{table:exhp}, and each model starts with the same low fidelity surrogate. 

\begin{figure}
\begin{center}
		\includegraphics[width=0.5\textwidth]{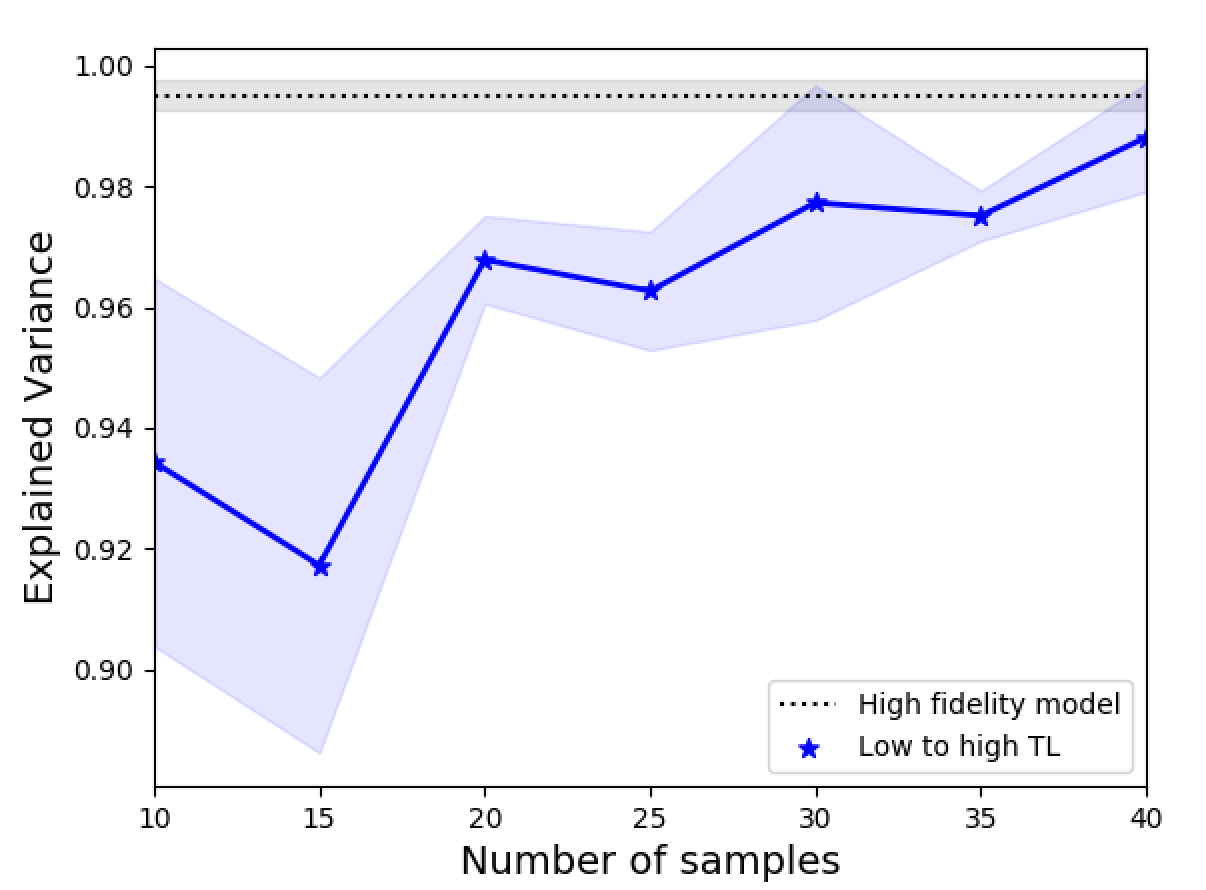}
		\caption{``High fidelity'' prediction quality as the number of high fidelity data points used for transfer learning from low fidelity data is increased. 30-40 high fidelity data points with transfer learning produce a model that is comparable in quality to one trained exclusively on 100 high fidelity simulations. }
		\label{fig:hf}
\end{center}
\end{figure}

\begin{figure}
\begin{center}
		\includegraphics[width=0.5\textwidth]{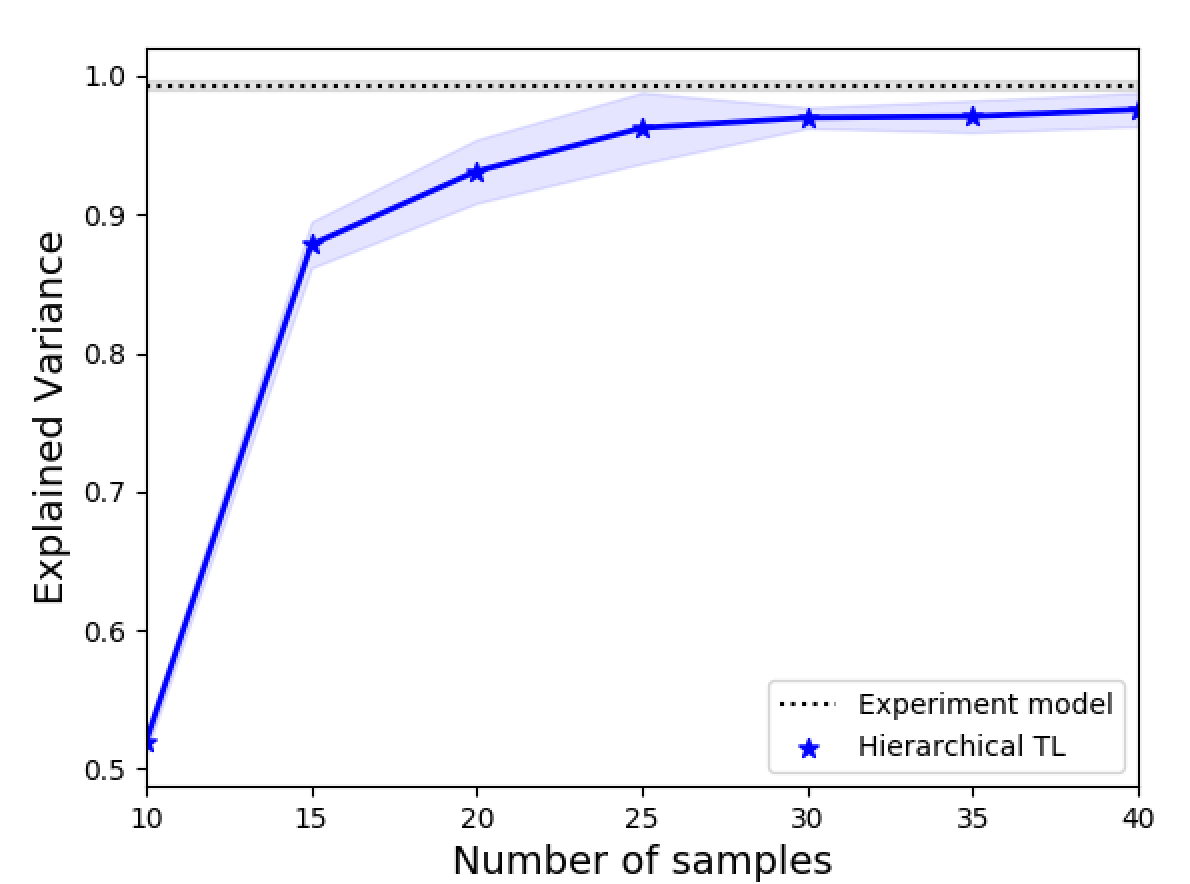}
		\caption{Experimental prediction accuracy as the number of experimental data points is increased in hierarchical transfer learning. Models are first trained on 100 low fidelity simulations, calibrated to high fidelity with 30 high fidelity data points, then subsequently calibrated to the experiments. The model quality converges with about 25 experiments, and is comparable to the performance of a model trained on 100 experiments alone.}
		\label{fig:hier}
\end{center}
\end{figure}

Figure \ref{fig:hf} illustrates how the transfer learning quality improves as the number of high fidelity data points is increased; the error bars reflect the variation in performance when the testing/training datasets are shuffled. Transfer learning with about 30 high fidelity samples produces models that perform similarly to models trained exclusively on high fidelity simulations. Thus the first step in the hierarchical calibration is performed with 30 high fidelity simulations. Next, we determine the minimum number of experimental data points needed to calibrate this model to the experiments. Figure \ref{fig:hier} illustrates the quality of the transfer learned model predictions as the number of experiments is varied. For this example, performance comparable to the baseline is achieved with 25 experiments; beyond this the improvement in model quality is minimal.

Whether it is beneficial to perform hierarchical transfer learning depends on the relative expense of the varying levels of fidelity in simulations and the experiments. If 100 low fidelity and 30 high fidelity simulations are less expensive than 100 high fidelity simulations, it is worth taking the hierarchical approach for subsequently calibrating to 25 experiments; however there may be situations in which running a high fidelity database is preferred. This example suggests that hierarchical transfer learning does offer improvements over transfer learning from low fidelity data directly to the experiments. The hierarchical calibration approach could also be improved by optimizing where the high fidelity simulations and experiments are placed in the design space; future work will explore sampling strategies for efficient hierarchical transfer learning. 

The previous results are for a very simple, low dimensional function that is fast to evaluate. In the next sections, we apply the same techniques to a real-world application: ICF experiments from the Omega laser facility.

\section{Transfer learning for ICF model calibration}\label{sec:tl2}
The performance of hierarchical transfer learning is next tested on a real-world dataset containing 23 experiments from the Omega laser facility that lie within a 9D design space. A database of 30k Latin hypercube sampled~\cite{lhs,doe} LILAC~\cite{lilac} simulations spans the 9D space encompassing the experiments. The nine input parameters varied in the databases include laser pulse parameters: the average drive, drive rise time, energy on the target, the first picket power, foot power, foot width, and foot picket width, and capsule geometry parameters: the ice thickness and the outer radius of the capsule. 
The 30k simulations are low fidelity; they are 1D, do not account of laser-plasma interactions (LPI) such as cross-beam energy transfer (CBET)~\cite{cbet,cbet1,cbet2}, and use flux-limited thermal diffusion models~\cite{fluxlimitdiff} . Each simulation takes about ten wall-clock minutes to run. Each of the 23 experiments is accompanied by a high fidelity simulation, which is a 1D LILAC simulation with CBET, more accurate equations of state~\cite{fpeos}, and nonlocal transport; these simulations require approximately ten wall-clock hours to run. Both the low and high fidelity simulations produce a large number of scalar outputs; 19 of which are included in the following analysis. There are only 5 observables available for all 23 experiments that are common to the simulation database that will be used to test transfer learning with experimental data. 

In section~\ref{sec:tl21}, we train the low fidelity simulation-based neural networks using DJINN. We will refer to the models trained only on the low fidelity LILAC simulations as ``low fidelity DJINN'' models. In section~\ref{sec:tl22}, we use transfer learning to calibrate from low fidelity to high fidelity simulations, and then to experiments.

\subsection{Low fidelity simulation-based DJINN models}\label{sec:tl21}
The low-fidelity simulation database is used to train an ensemble of five low-fidelity DJINN models, which predict all 19 observables simultaneously and are individually cross-validated. The variance between DJINN models, each of which have been trained on a different random 80\% subset of the data, will provide uncertainty estimates on the model predictions. The hyper-parameters used to train the networks are summarized in Table \ref{table:omegahp}. The mean explained variance score of the low fidelity models for each output is given in Table \ref{table:lilacEV}.

\begin{table}[]
\begin{center}
\caption{Hyper-parameters for original model and transfer learning for the Omega dataset.}
\label{table:omegahp}
\begin{tabular}{ll}
\hline
\multicolumn{2}{c}{Original Model Parameters} \\ \hline
\multicolumn{1}{l|}{Number of models} & 5 \\
\multicolumn{1}{l|}{Hidden layer widths} & \begin{tabular}[c]{@{}l@{}}11-13-22-16; 11-14-19-26; 11-14-24-16;\\ 11-15-29-30; 11-14-22-29\end{tabular} \\
\multicolumn{1}{l|}{Learning rate} & 0.004 \\
\multicolumn{1}{l|}{Batch size} & 1500 \\
\multicolumn{1}{l|}{Epochs} & 400 \\ \hline
\multicolumn{2}{c}{Transfer Learning Parameters} \\ \hline
\multicolumn{1}{l|}{Retrained layers} & Final 2 \\
\multicolumn{1}{l|}{Learning rate} & 0.0003 \\
\multicolumn{1}{l|}{Batch size} & 1 \\
\multicolumn{1}{l|}{Epochs} & 2300
\end{tabular}
\end{center}
\end{table}

\begin{table*}[]
\begin{center}
\caption{Mean explained variance scores on the test datasets for five DJINN models trained on random 80\% subsets of the 30k low-fidelity LILAC simulations.}
\label{table:lilacEV}
\begin{tabular}{lll}
\hline
AbsorptionFraction:\,\, 0.991 & PeakKineticEnergy:\,\, 0.981 & Tion$\_$DD:\,\, 0.959  \\
Adiabat:\,\, 0.886            & Pressure:\,\, 0.958          & Vi: \,\,0.971           \\
BW: \,\,0.869                 & R0: \,\,0.962                & Yield: \,\,0.944       \\
BangTime:\,\, 0.990           & RhoR: \,\,0.968              & Yield$\_$DD:\,\, 0.949 \\
ConvergenceInner:\,\, 0.964   & ShockMass:\,\, 0.856          & rhonave: \,\,0.967 \\
ConvergenceOuter:\,\, 0.962    & rhomaxbt: \,\,0.962          &   \\
IFAR: \,\,0.885               & Tion: \,\,0.956              &       \\            \hline
\end{tabular}
\end{center}
\end{table*}

The same set of low fidelity DJINN models are used for standard (low fidelity simulations to experiments) and hierarchical transfer learning (low to high fidelity simulations to experiments). 

\subsection{Hierarchical transfer learning with the Omega dataset}\label{sec:tl22}
The hierarchical approach, in which the low fidelity model is calibrated first to high fidelity simulation data, then to the experimental data, is compared to transfer learning directly from low fidelity simulations to experiments for the Omega database. The models transfer learned to high fidelity simulations are referred to as ``high fidelity DJINN'' models, and those that are subsequently transfer learned to experiments as ``experiment DJINN'' models. If the high fidelity simulations are more accurate depictions of reality than the low fidelity simulations, priming the DJINN model with high fidelity information prior to calibrating to the experiments could improve the ability of the model to adapt to the experimental data.  

The low fidelity DJINN models described in the previous section are calibrated independently, each on a different random subset of the high fidelity or experimental data. For each of the models, the first three layers of weights are frozen, and the remaining two layers are available for retraining, as shown in the cartoon in Figure \ref{fig:tl2}. Note that the architecture of the networks is not reflected in this cartoon; the true architectures are given in Table \ref{table:arch} for the ensemble of five DJINN models. 
The last two layers of weights are retrained to convergence for 2000 epochs with a batch size of one and a learning rate of 0.0003 in the Adam optimizer. Each model is trained on a random 90\% of the experimental data (20 points) and tested on the remaining 10\% (3 points). The low fidelity and post-shot simulations have 19 outputs, but the experiments only have 5 available observables. To calibrate to the experiments, the cost function, which is the MSE of the 19 scaled outputs, is modified such that the missing 14 outputs are not weighted. More explicitly, the cost becomes a weighted MSE where the weights are zero for outputs not measured in the experiment, and unity for those that are observed in the experiment. The predictions for the remaining 14 observables may change in non-physically motivated ways, thus we will focus only on the 5 available observables. An equivalent approach to the weighted cost function would be to train the low fidelity and high fidelity DJINN models with only the 5 outputs available in the experiment. We choose to retain all 19 outputs so we can evaluate the accuracy of the high fidelity calibration for all 19 observables in the hierarchical modeling approach. Note that all input and output data is scaled [0,1], using the parameter ranges set by the database of 30k simulations, prior to training. This prevents the cost function from being biased toward outputs that are larger in magnitude due to the choice of units.

\begin{table}[]
\centering
\caption{Architectures of DJINN models trained on the Omega databases. There are nine inputs and nineteen outputs for the baseline, low fidelity models.}\label{table:arch}
\begin{tabular}{l}
\hline
Architectures           \\ \hline
(9, 10, 13, 20, 20, 19) \\
(9, 10, 11, 23, 21, 19) \\
(9, 11, 11, 13, 20, 19) \\
(9, 11, 12, 19, 25, 19) \\
(9, 10, 13, 22, 25, 19) \\ \hline
\end{tabular}
\end{table}

\begin{figure*}
\begin{center}
		\includegraphics[width=0.95\textwidth]{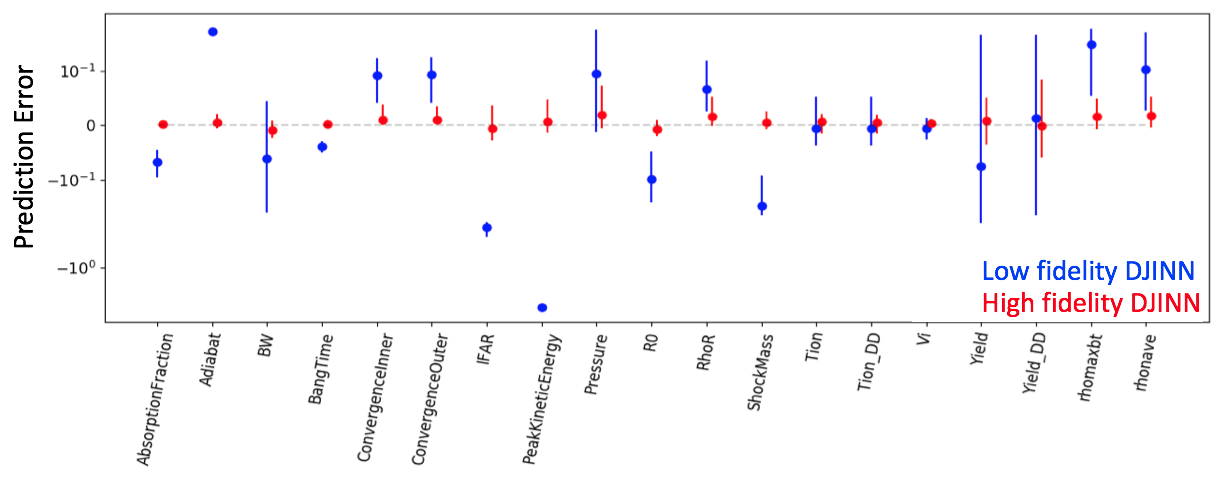}
		\caption{Prediction error (with the high fidelity simulation as the ground truth) for calibrated and uncalibrated DJINN models. The low fidelity model predicts high fidelity observables with significant error, as the models contain different physics. The models calibrated to high fidelity data are able to predict all 19 high fidelity observables with low error.}
		\label{fig:postshot}
\end{center}
\end{figure*}

First we consider transfer learning from the low fidelity simulations to the high fidelity simulations. Figure~\ref{fig:postshot} illustrates the prediction error (calculated on training and testing data combined), computed as: 

\begin{equation}
\mathrm{Error = \frac{(Prediction) - (High\, fidelity\, truth)}{(High\, fidelity\, truth)}},
\end{equation}

for all nineteen available outputs for the uncalibrated (low fidelity) and calibrated (high fidelity) DJINN models. The error bars reflect the standard deviation in prediction errors from the ensemble of DJINN models; the points on Fig.~\ref{fig:postshot} illustrate the mean error. 

The low fidelity and high fidelity simulations differ significantly in their predictions of the nineteen observables, shown by the error in the blue points of Figure~\ref{fig:postshot}. The low error in the red points indicates that the networks are able to successfully learn the high fidelity outputs via transfer learning.

The high fidelity calibrated models are next calibrated to the experimental data, again by transfer learning the last two layers of the network -- the same layers that were modified to calibrate to the high fidelity data. The mean and standard deviation of final prediction error is now computed using the experiment as the ground truth:

\begin{equation}
\mathrm{Error = \frac{(Prediction) - (Experiment\, truth)}{(Experiment\, truth)}},
\end{equation}

\begin{figure*}
\begin{center}
		\includegraphics[width=0.85\textwidth]{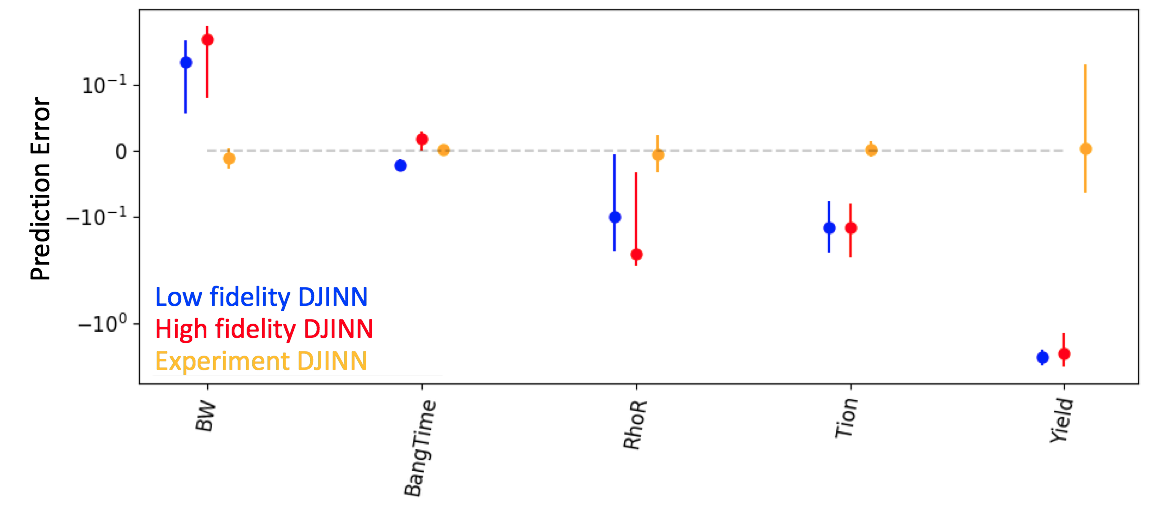}
		\caption{Prediction error (with the experiment as the ground truth) for low fidelity DJINN models, DJINN models calibrated to the high fidelity data, and DJINN models calibrated to the experimental data. The experimentally calibrated models predict the five experimental observables with low error.}
		\label{fig:exp}
\end{center}
\end{figure*}

The prediction quality is illustrated in Fig.~\ref{fig:exp} for the low fidelity, high fidelity, and experiment DJINN models. 

Figure \ref{fig:exp} illustrates that the high fidelity simulations are not necessarily more predictive of reality than the low fidelity simulations, however transfer learning to experiments is still successful. The largest error for transfer learning is in the yield, due to the model needing to adjust its predictions by over an order of magnitude for most experiments, however the mean prediction error for all observables is close to zero. 

Since the post-shots are no closer to the experiments than the low fidelity models, hierarchical modeling does not offer significant benefits for this dataset; comparable results are achieved by transfer learning directly from low fidelity simulations to the experiments. Table \ref{table:omega} records the average explained variance ratio for each the experimental observables for the hierarchical models, and those that are calibrated directly from low fidelity to experiments. The explained variance ratios are computed on the test dataset, and are averaged for the five models.

\begin{table}[]
\centering
\caption{Explained variance scores for models calibrated from low fidelity to high fidelity simulations to experimental data, and for models calibrated directly from low fidelity simulations to experiments. The high fidelity simulations are not an accurate picture of reality, and thus there are no significant benefits of first calibrating to the high fidelity data for this particular dataset.}
\label{table:omega}
\begin{tabular}{l|l|l|l}
           & \multicolumn{2}{l|}{Mean$\pm$SD Explained Variance} &         \\ \hline
Observable & Hierarchical               & Low fi. - Exp.           & p-value \\ \hline
Burnwidth  & 0.975 $\pm$ 0.023          & 0.889 $\pm$ 0.079         & 0.139   \\
Bangtime   & 0.987 $\pm$ 0.015          & 0.942 $\pm$ 0.092         & 0.364   \\
$\rho$R    & 0.874 $\pm$ 0.097          & 0.835 $\pm$ 0.179         & 0.712   \\
T$_{ion}$  & 0.988 $\pm$ 0.009          & 0.924 $\pm$ 0.094         & 0.211   \\
Yield      & 0.818 $\pm$ 0.143          & 0.956 $\pm$ 0.034         & 0.096   \\ \hline
\end{tabular}
\end{table}

\begin{figure*}
\begin{center}
		\includegraphics[width=0.95\textwidth]{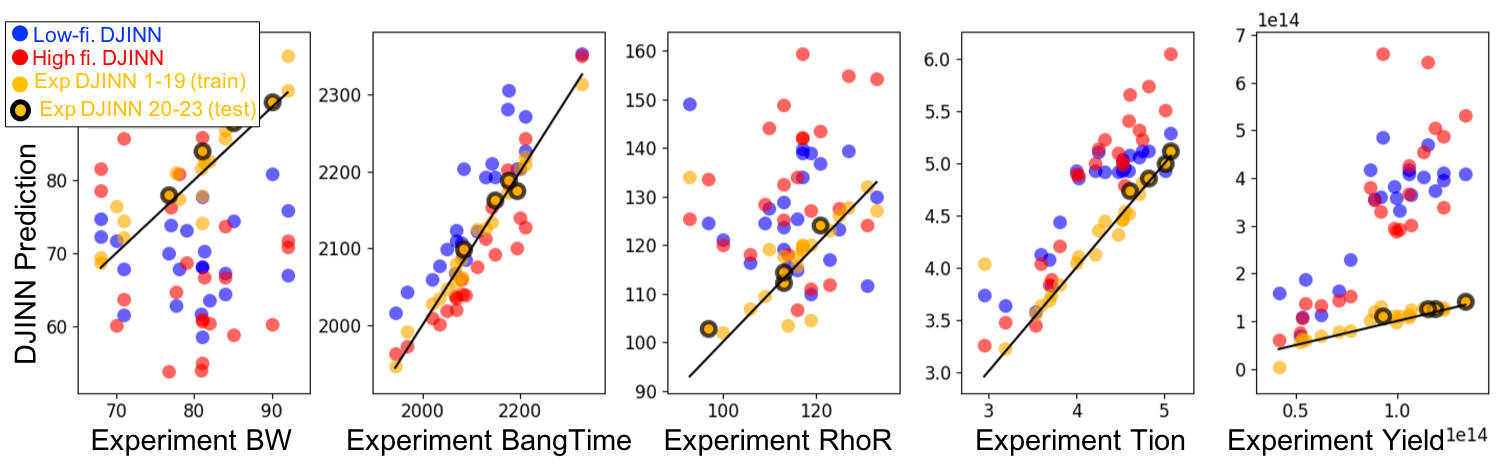}
		\caption{Predictions of new experiments (yellow circled in bold black) after transfer learning using previous experimental data. The models are able to accurately predict new experiments, which are small extrapolations in input space from the old experimental data. The yellow data points indicate the training data predictions, the points with black outlines are predictions for the test data.}
		\label{fig:futurepred}
\end{center}
\end{figure*}

The previous analyses always involve randomly choosing the training and testing data for model calibration. To illustrate how these models can be used to predict the outcomes of future experiments, we take the models calibrated to the high fidelity data and calibrate to experiments using only the oldest 19 experiments in the dataset. We then test the models on the 4 most recent experiments; the predictions are shown in Figure \ref{fig:futurepred}. Training on the old data and predicting the four most recent experiments requires the model to extrapolate in input space, away from the old experimental data. The model is able to successfully predict the outcomes of the newest four experiments, demonstrating it does have the ability to successfully extrapolate away from the experimental data, but within the bounds of the simulations.

\section{Exploring discrepancies between simulations and Omega experiments with transfer learning}\label{sec:tl3}
A result of hierarchical transfer learning is that we now have models that emulate low fidelity simulations, high fidelity simulations, and experiments. We can use these three models to explore the 9D design space and study the discrepancies between the two types of simulations and the experiments. 

\begin{figure*}
\begin{center}
		\includegraphics[width=0.75\textwidth]{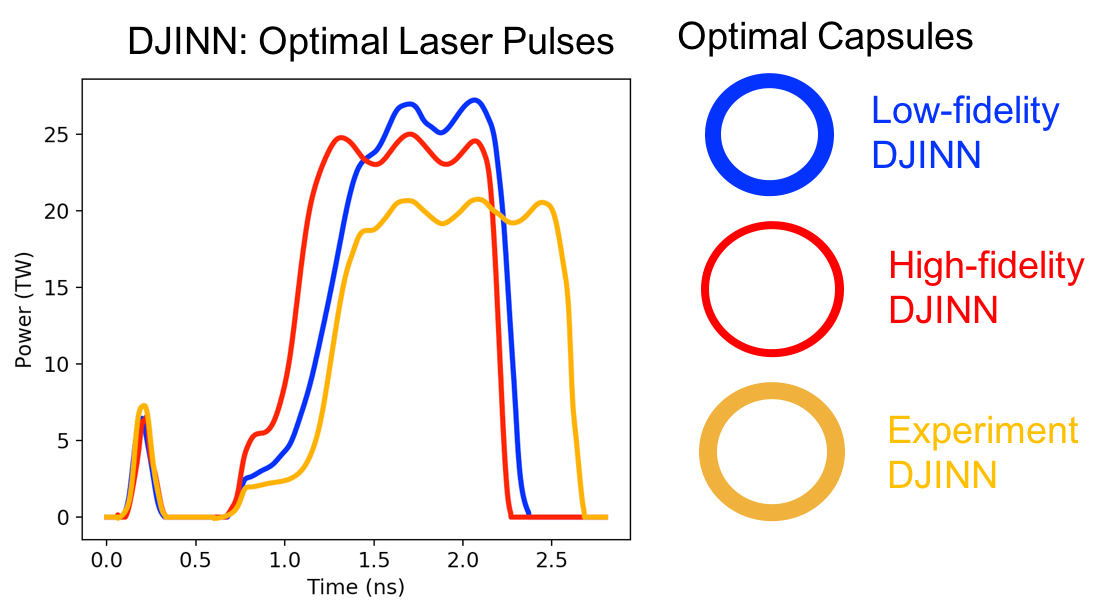}
		\caption{Designs which optimize ITFX according to the low fidelity, post-shot, and experiment DJINN models. The three designs are distinct due to the lack of accurate physics models, asymmetries, and other experimental sources of performance degradation not included in the simulations.}
		\label{fig:designs}
\end{center}
\end{figure*}

A primary use of ICF implosion simulations is to find optimal design settings for experiments. An interesting application of the three models is thus to search for ``optimal'' designs using each fidelity surrogate and determine if the simulation-based models suggest a similar ``optimal'' design as the experiment-informed model. 
For this exercise, we define an optimal design as one that maximizes the experimental ignition threshold factor (ITFX)~\cite{ignition_metric2}:

\begin{equation}
\mathrm{ITFX \propto Yield \cdot (\rho R)^2}.
\end{equation}

where $\rho R$ is the areal density. The resulting optimal designs are illustrated in Fig.~\ref{fig:designs}. 

There are several important differences between the three optimal designs. First, consider the differences between the low fidelity (blue) and high fidelity (red) designs. The low fidelity design prefers high compression of a thick capsule, and achieves this by driving the capsule with a very high power. This differs from the high fidelity design, which prefers a lower power and thinner shell in order to achieve a similar implosion velocity and yield. The high fidelity design includes CBET effects, so it is reasonable for this design to lower the peak power to avoid CBET. 

Next consider the experimental design: unlike the high fidelity design, this design adjusts the picket and the foot of the pulse to guard against hydrodynamic instabilities that are not modeled in 1D simulations. Lowering the foot of the pulse (which occurs around 0.75 ns) lowers the adiabat inside the shell, allowing for higher compression and therefore higher areal density. To mitigate instabilities associated with higher compression, the picket of the pulse (occurring around 0.2 ns) is increased to increase the adiabat on the outer surface of the shell. The higher outer adiabat reduces hydrodynamic instabilities by increasing the ablation velocity~\cite{Goncharov:2003ss, Knauer:2005cq, Anderson:2004hh}. The experimental capsule is thicker and is driven at an even lower power than the post shot for a longer period of time; perhaps due to underestimation of the CBET effects by the high fidelity simulations. 

The maximum ITFX design according to the experiment-calibrated model is consistent with other analyses of this database~\cite{varchasaps,bettiaps}. The researchers at Omega have been training power law-based models to relate simulation outputs and experimental measurements; through this process they found that to optimize yield, they should increase the thickness of the ice in the capsule, as suggested by the experimentally-calibrated DJINN model. They confirm their model predictions with a series of experiments, each time making small extrapolations in shell thickness and updating their models with the experimental before predicting the outcome of the next experiment. After maximizing the yield, the researchers performed a set of experiments to independently optimize the areal density by modifying the picket and foot of the pulse. This approach is largely physics-guided, and treats the pulse and capsule independently to optimize yield and areal density; it does not explicitly account for interactions between the capsule geometry and the laser pulse.  

The neural network-based optimization can consider nonlinear interactions between the inputs, tuning the pulse and capsule simultaneously to maximize ITFX. However, the neural networks might be inaccurate far from the experimental data, thus caution should be taken to make small extrapolations from the data with this technique as well. The fact that two distinct methods for creating data-driven models for the Omega database suggest similar design choices for optimizing performance is encouraging, and illustrates the powerful role transfer learning can play in creating predictive models of ICF experiments.

\section{Conclusions}
Transfer learning is the process of taking a neural network trained on a task for which there are copious amounts of data (such as low fidelity simulations), freezing many layers of the network, and retraining the unfrozen layers on a small set of expensive data (such as high fidelity simulations or experiments). This method enables the creation of neural networks which emulate expensive simulations or experiments without requiring massive quantities of expensive data.

In this work, we introduce the idea of hierarchical transfer learning, the process of calibrating from low to high fidelity simulations to experiments, which enables the creation of accurate emulators for the experiment with minimal computational cost. We apply hierarchical transfer learning to a collection of ICF simulations and experiments from the Omega laser facility, and demonstrate the ability to create models which are more predictive than simulations alone. Hierarchical transfer learning offers a promising framework for integrating various fidelity simulations and experiments into a common predictive framework that can be used for data-drive design optimization and to study the discrepancies between simulations and experiments. 

\section{Acknowledgments}
The authors would like to thank Riccardo Betti, Varchas Gopalaswamy, and Jim Knauer at the Laboratory for Laser Energetics for providing the LILAC simulations, the Omega experimental databases, and for candid conversation. 

This work was performed under the auspices of the U.S. Department of Energy by Lawrence Livermore National Laboratory under Contract DE-AC52-07NA27344. Released as LLNL-CONF-764063.
This document was prepared as an account of work sponsored by an agency of the United States government. Neither the United States government nor Lawrence Livermore National Security, LLC, nor any of their employees makes any warranty, expressed or implied, or assumes any legal liability or responsibility for the accuracy, completeness, or usefulness of any information, apparatus, product, or process disclosed, or represents that its use would not infringe privately owned rights. Reference herein to any specific commercial product, process, or service by trade name, trademark, manufacturer, or otherwise does not necessarily constitute or imply its endorsement, recommendation, or favoring by the United States government or Lawrence Livermore National Security, LLC. The views and opinions of authors expressed herein do not necessarily state or reflect those of the United States government or Lawrence Livermore National Security, LLC, and shall not be used for advertising or product endorsement purposes.

\bibliographystyle{IEEEtran}
\bibliography{bib.bib}

\end{document}